\documentclass[final]{cvpr}

\usepackage{times}
\usepackage{epsfig}
\usepackage{graphicx}
\usepackage{subfigure}
\usepackage{amsmath}
\usepackage{amssymb}
\usepackage{ragged2e}
\usepackage{multirow}
\usepackage{booktabs}

\usepackage[pagebackref=true,breaklinks=true,colorlinks,bookmarks=false]{hyperref}

\newlength\savewidth\newcommand\shline{\noalign{\global\savewidth\arrayrulewidth
  \global\arrayrulewidth 1pt}\hline\noalign{\global\arrayrulewidth\savewidth}}

\renewcommand{\raggedright}{\leftskip=0pt \rightskip=0pt plus 0cm}
\def\cvprPaperID{880} 
\def\confYear{CVPR 2021}

\begin{document}
\title{\ Learning Salient Boundary Feature for Anchor-free \\ Temporal Action Localization}

\author{
Chuming Lin$^{*}$\textsuperscript{\rm 1},
Chengming Xu$^{*}$\textsuperscript{\rm 2},
Donghao Luo\textsuperscript{\rm 1},
Yabiao Wang\textsuperscript{\rm 1},
Ying Tai\textsuperscript{\rm 1}, \\
Chengjie Wang\textsuperscript{\rm 1},
Jilin Li\textsuperscript{\rm 1},
Feiyue Huang\textsuperscript{\rm 1},
Yanwei Fu\textsuperscript{\rm 2}\\
\textsuperscript{\rm 1}Youtu Lab, Tencent,
\textsuperscript{\rm 2}Fudan University, China\\
{\tt\small \{chuminglin, michaelluo, caseywang, yingtai, jasoncjwang, jerolinli, garyhuang\}@tencent.com} \\
{\tt\small \{cmxu18, yanweifu\}@fudan.edu.cn } \\
}

\maketitle

\let\thefootnote\relax\footnotetext{$^{*}$ indicates equal contributions.}
\let\thefootnote\relax\footnotetext{This work was done when Chengming Xu was an intern at Tencent Youtu Lab. Yanwei Fu is the corresponding author.}
\begin{abstract}
    Temporal action localization is an important yet challenging task in video understanding. Typically, such a task aims at inferring both the action category and localization of the start and end frame for each action instance in a long, untrimmed video.
    While most current models achieve good results by using pre-defined anchors and numerous actionness, such methods could be bothered with both large number of outputs and heavy tuning of locations and sizes corresponding to different anchors. Instead, anchor-free methods is lighter, getting rid of redundant hyper-parameters, but gains few attention. In this paper, we propose the first purely anchor-free temporal localization method, which is both efficient and effective. Our model includes (i) an end-to-end trainable basic predictor, (ii) a 
   saliency-based refinement module to gather more valuable boundary features for each proposal with a novel boundary pooling, and (iii) several consistency constraints to make sure our model can find the accurate boundary given arbitrary proposals. Extensive experiments show that our method beats all anchor-based and actionness-guided methods with a remarkable margin on THUMOS14, achieving state-of-the-art results, and comparable ones on ActivityNet v1.3. Code is available at \url{https://github.com/TencentYoutuResearch/ActionDetection-AFSD}.
\end{abstract}

\section{Introduction \label{sec:1}}

Recently, with the progress of technology, a dramatically increasing number of videos have been stored and accessible from various daily activities. Temporal Action Localization (TAL), as a fundamental aspect of video understanding, thus plays an important role in real life, extending in several practical applications such as video analysis and summarization, human interaction, \textit{etc}. Compared with action recognition that takes medium-range videos as input and only requires class labels as prediction, TAL is aimed at not only classifying every activity instance in each video, but also looking for the accurate temporal locations of them.

\begin{figure}[t]
    \centering
    \includegraphics[scale=0.85]{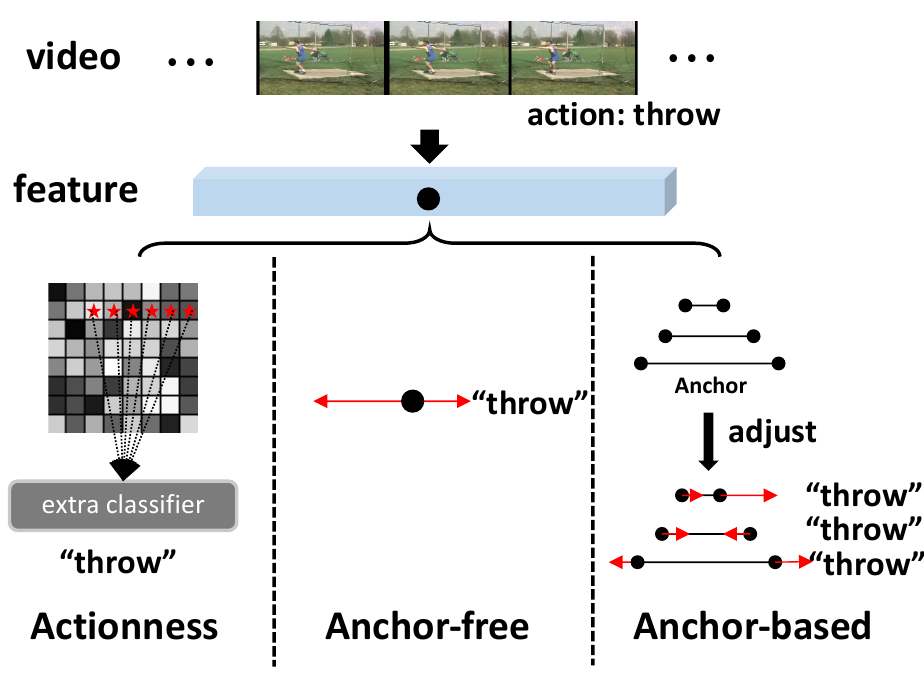}
    \caption{Compared with actionness and anchor-based methods, anchor-free method is more efficient and flexible to produce fewer proposals without any extra classifier and pre-defined anchors.}
    \label{fig:intro_1}
    \vspace{-0.15in}
\end{figure}

Current TAL models mainly focus on learning actionness of each frame~\cite{lin2018bsn, lin2019bmn, lin2020dbg, luo2020weakly, ma2020sf, long2020learning} or adjustment of pre-defined anchors~\cite{xu2017rc3d, long2019gaussian, liu2020progressive}, named actionness-guided methods and anchor-based methods, as shown in Fig.~\ref{fig:intro_1}. In spite of reasonable good results on benchmark datasets, such methods are still limited to the following points: (1) Both methods will produce a bunch of redundant proposals. For example, given a video with $T$ frames, we have to produce 
$\mathcal{O}(T^2)$ and $C\cdot T$ proposals for the ``actionness-guided'' BSN~\cite{lin2018bsn}, and ``anchor-based'' R-C3D~\cite{xu2017rc3d}, individually. Here  $C$ is the number of pre-defined anchors.
 These proposals  lead to prohibitive computational cost in both calculating the training loss and post-processing for testing.
(2)
Actionness-guided methods can solely provide predictions of temporal boundary, while they have to rely on the extra  model such as S-CNN~\cite{shou2016scnn} and P-GCN~\cite{zeng2019pgcn} for classification. Nevertheless, the models of two stages are isolated and thus incapable of sharing information for the end-to-end update.
 (3) Typically,
anchor-based methods are very sensitive to some critical hyper-parameters, such as the number and size of pre-defined anchors; and it is very non-trivial to tune these hyper-parameters in the real-world applications.

Alternatively, an efficient recipe for localization is to resort to the \textit{anchor-free} method, which does not require pre-defined anchors.
Typically, this type of method only generates one proposal for each temporal location in the form of a pair of values representing the distance between the start and end moments to the current location, individually.

In contrast to the existed methods, anchor-free model saves huge amount of pre-defined anchors, while assembling both boundary regression and classification in one model, thus being productive. Furthermore, even though some pilot studies, \textit{e.g.},~Yang \textit{et. al.}~\cite{yang2020a2net} observed relatively weak results for anchor-free methods, the supporting evidence in object detection~\cite{zhang2020bridging} shows that
such methods with well designed network structure and training strategy should, in principle, be comparable with anchor-based ones.

\begin{figure}[t]
    \centering
    \includegraphics[scale=0.4]{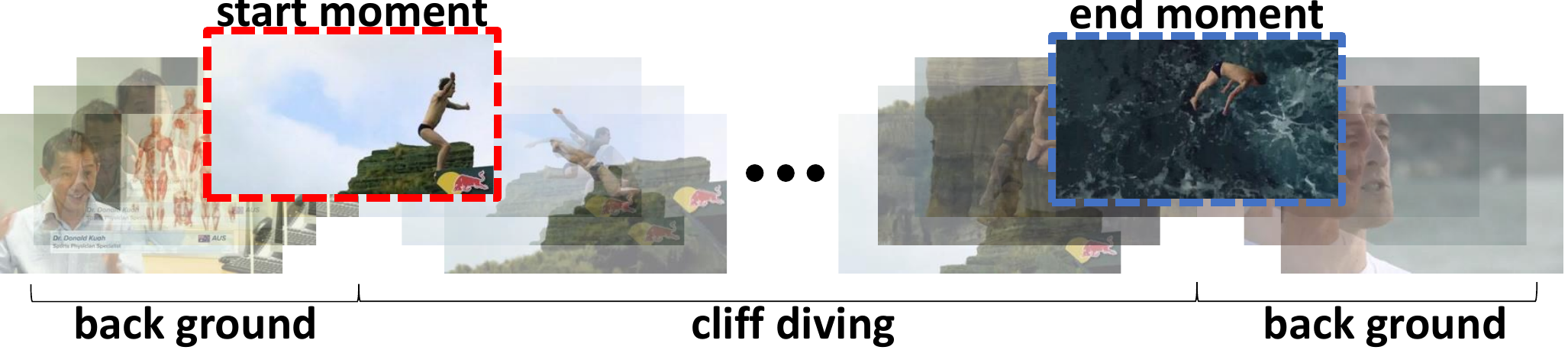}
    \caption{An action instance of cliff diving. Note that the start and end moments of the movement are more salient than others, which can bring us significant information to judge the boundary and completeness of the action.}
    \label{fig:intro_2}
    \vspace{-0.15in}
\end{figure}

To this end, in this paper we propose a novel purely anchor-free TAL framework dubbed Anchor-Free Saliency-based Detector (AFSD). Essentially, we first build a naive anchor-free predictor containing an end-to-end trainable backbone network, a feature pyramid network and a simple prediction network which outputs the action class and the temporal distance of the start and end from each location. To learn a more accurate boundary, we refer to former TAL methods~\cite{lin2018bsn, lin2019bmn} indicating the importance of boundary or context feature. These works obtain such features mainly by merging the neighbor of start and end moments with convolutions or mean pooling. However, we claim that in fact moment-level feature is more valuable than region-level feature for distinguishing whether an action starts or ends. As shown in Fig.~\ref{fig:intro_2}, the background regions near the start and end moments are showing other irrelevant scenes, while regions inside the action are almost the same, which cannot provide any information for judging if the action starts or ends. Such an example indicates the importance of a moment-level feature.

Therefore, we propose a novel boundary pooling which, instead of aggregating the whole region, tries to find the most salient moment-level feature for both start and end regions. We further equip the boundary pooling with a newly proposed Boundary Consistency Learning (BCL) strategy to regularize the pooling operation to provide the correct boundary features for each action. In detail, we employ a modified ground truth signal indicating start and end moments to guide the model. Then, we rearrange the video clips to help model discriminate background and action features by self-supervised contrastive learning. We conduct extensive experiments on THUMOS14 and ActivityNet1.3. On THUMOS14 our model attains 3.7\% improvement on \textit{m}AP@0.5 against the state-of-the-art methods. The results on ActivityNet1.3 are also comparable.

In summary, our paper has the following contributions:
\begin{enumerate}
    \item We, for the first time, propose a purely anchor-free temporal action localization model. This model enjoys not only less hyper-parameter to tune and less outputs to process, but also better performance, thus making the best of both worlds.
    \item To make full use of the anchor-free framework, we discuss the impact of boundary features and propose novel boundary pooling method whose output is used along with the coarse proposals to generate fine-grained predictions. Moreover, we introduce a novel Boundary Consistency Learning strategy which can constrain the model to learn better boundary feature.
\end{enumerate}

\section{Related Work}
\noindent \textbf{Anchor-based Localization}
Anchor-based localization models rely on adjusting the pre-defined anchors. 
TURN~\cite{gao2017turn} aggregates features from basic video unit for clip-level features, which are used to classify the activity and regress the temporal boundary. R-C3D~\cite{xu2017rc3d} takes the inspiration from faster-RCNN~\cite{ren2015faster}, which utilizes a streamline including proposal generation, proposal-wise pooling and final prediction. GTAN~\cite{long2019gaussian} modifies the pooling procedure, adopting a weighted average via a learnable Gaussian kernel for each proposal. Due to the fixed pre-defined anchors, such methods are not flexible when it comes to various action classes. Different from them, our model does not require tuning extra hyper-parameters for anchors, thus more efficient.

\noindent \textbf{Actionness-guided Localization}
Unlike anchor-based methods, actionness-guided methods mainly focus on evaluating `actionness', which indicates the probability of a potential action, for each frame or clip in a video. The actionness is afterwards post-processed to generate action proposals. Zhao \textit{et. al.} designed SSN~\cite{zhao2017ssn} in which course proposals are first divided into three semantic parts, learned respectively. Next probability of activity and completeness is predicted and used to merge different proposals. Lin \textit{et. al.} proposed BSN~\cite{lin2018bsn} which learns to predict start, end and actionness of each temporal location. The proposals are generated by gathering locations with high start and end probabilities, with low confidence ones further abandoned by an evaluation module. They later improved this framework to BMN~\cite{lin2019bmn}, which additionally generates a Boundary-Matching confidence map to help get better proposals.
While no pre-defined anchors are required for actionness-guided methods, such methods are more like enumeration methods where all possible combinations of temporal locations are considered, thus totally different from anchor-free localization where boundaries are directly predicted for each time step.

\noindent \textbf{Anchor-Free Object Detection}
Analogous to TAL, there is a surge of the usage of anchor-free methods in object detection. YOLO~\cite{redmon2016yolo} is the most well-known anchor-free method, in which a neural network model is directly used to predict coordinates of bounding boxes from raw images. Such framework is too simple, thus suffering from poor performance. Following works mainly focus on improving performance via setting different prediction targets and using more detailed features. CornerNet~\cite{law2018cornernet} lets the model learn to predict top-left and bottom-right keypoints of each bounding box. FCOS~\cite{tian2019fcos} aims to learn the distance to boundaries of each spatial location and utilizes feature pyramid for objects with diverse scales. BorderDet~\cite{qiu2020borderdet} modifies the RoI pooling into BorderAlign to get more powerful proposal-level feature. We take inspiration from these methods to design a basic anchor-free localizer, along with making full use of the temporal insights of videos to propose novel refinement strategy and consistency learning.

\noindent \textbf{Contrastive Learning}
There has been an increasing interest in contrastive learning used in unsupervised learning. Compared with the application of contrastive learning in image understanding~\cite{chen2020simclr, he2020moco}, fewer contributions of contrastive learning have been made in video understanding. Guillaume \textit{et. al.}~\cite{lorre2020temporal} proposed a temporal contrastive training strategy for action recognition, in which an autoregressive model is used to predict future video segment given enough information, and then compare the prediction with ground truth. Gong \textit{et. al.}~\cite{gong2020learning} adopted a contrastive scoring method to evaluate action proposals in unsupervised temporal localization in the inference phase. Compared with these works, we make further trial into leveraging contrastive learning to help train a supervised temporal localization model, which has never been studied before.

\section{Method}
Denote a video dataset as $\mathcal{T}=\{\mathcal{T}^{train}, \mathcal{T}^{test}\}$, each data instance $\{X, \Psi_X\}$ contains a video $X=\{x_t\}_{t=1}^T$ with $T$ RGB frames or optical flows. The corresponding annotation $\Psi_X$ can be depicted as tuples
$\{(\phi_m, y_m)\}_{m=1}^{M_X}$ where $M_X$ is the number of action instances in $X$, $\phi_m=(\psi_m, \xi_m)$ denotes the start time, end time and $y_m$ indicates the action category. Our goal is to train a model to predict proposals with class scores which could have high recall and precision with the ground truth on the test set $\mathcal{T}^{test}$.

\begin{figure*}
    \centering
    \includegraphics[scale=0.73]{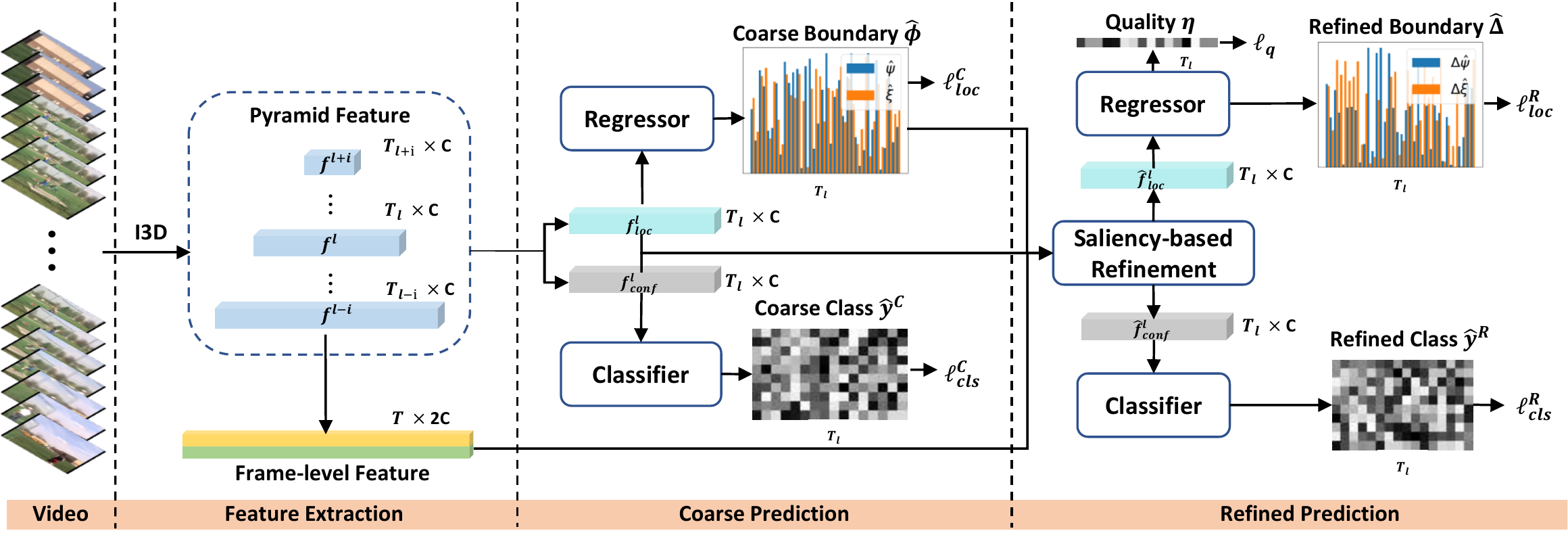}
    \caption{The overview of our approach. Given an input video $X$, we employ I3D model to extract feature and construct 1D temporal pyramid features. Next, each pyramid feature is utilized to generate coarse proposals via basic prediction module. Finally, our saliency-based refinement module will adjust the class score, start and end boundaries and predict the corresponding quality score for each coarse proposal. Note that our model is a fully end-to-end method and trained with I3D feature extraction network without any preprocessing.}
    \label{fig:overview}
    \vspace{-0.1in}
\end{figure*}

\begin{figure}
    \centering
    \includegraphics[scale=0.5]{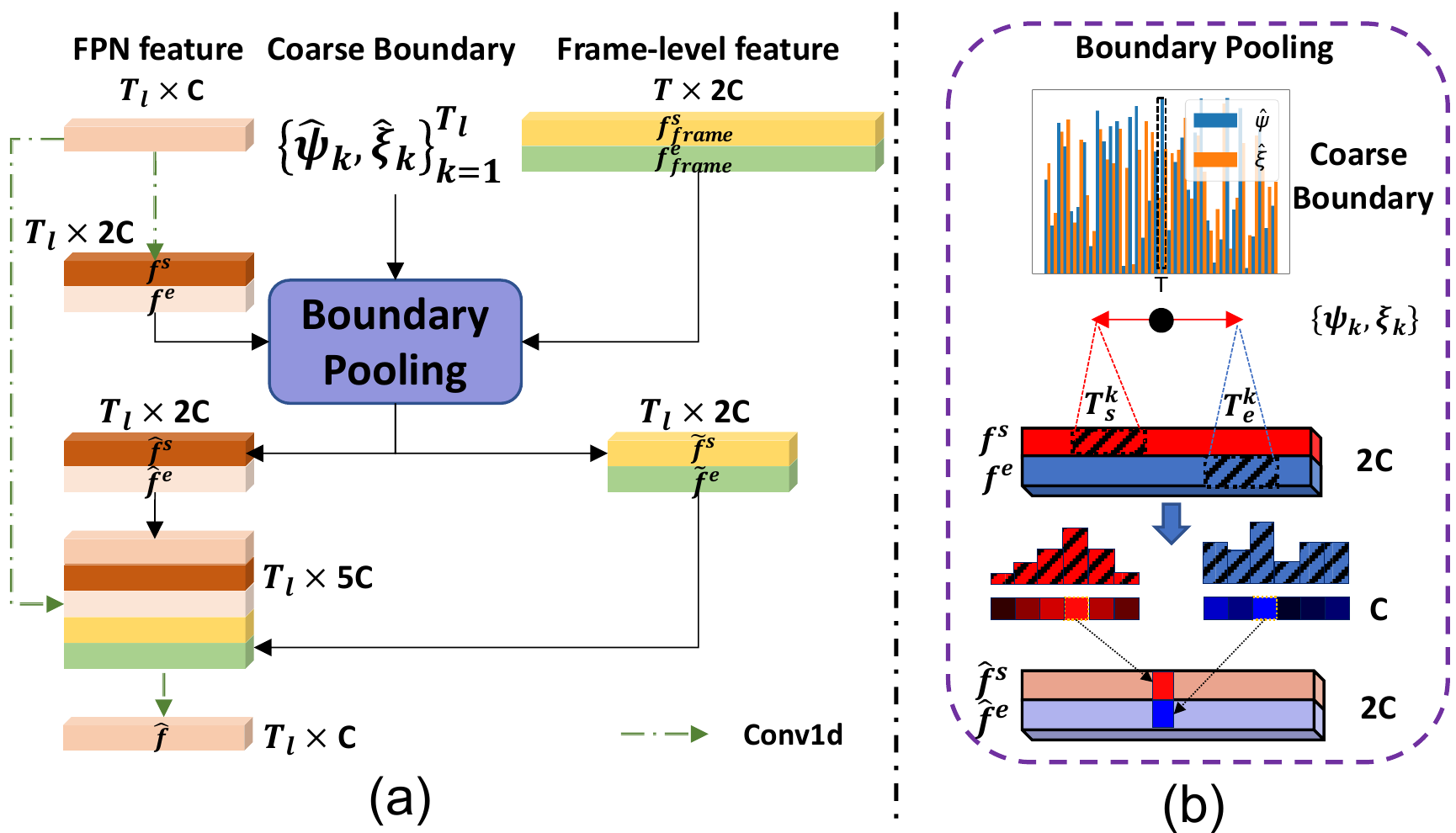}
    \caption{(a) Saliency-based Refinement Module: utilize coarse boundaries, FPN feature and frame-level feature to construct salient boundary feature. (b) Boundary Pooling: search salient moment features in the boundary regions of the input feature.}
    \label{fig:refinement}
    \vspace{-0.15in}
\end{figure}

\noindent \textbf{Overview} We propose a purely anchor-free architecture named AFSD which is shown in Fig.~\ref{fig:overview}. Concretely, given a video $X$, we first process the video with a backbone network and a feature pyramid network. Take RGB frames as example, for each video $X$, we use a Kinetics pre-trained I3D~\cite{carreira2017quo} model to extract a 3D feature $F\in \mathbb{R}^{T^{'}\times C^{'}\times H^{'}\times W^{'}}$, where $T^{'},C^{'}, H^{'}, W^{'}$ denote the time step, channel, height and width individually. This feature is afterwards flattened along the last three dimensions to a 1D feature sequence. Such a sequence can contain the temporal and spatial information of whole video. We then exert a feature pyramid network including several temporal convolutions, of which the detailed architecture is shown in our supplement, to merge the spatial dimension and aggregate the temporal dimension in different levels. The pyramid features are further utilized to generate a coarse proposal sequence $\{(\hat{\psi}_i, \hat{\xi}_i, \hat{y}_i^C)\}$ with a basic anchor-free prediction module (Sec.~\ref{sec:basic}), which includes a simple regressor and classifier. After that for each proposal, the predicted temporal regions are employed to get the salient boundary features with the boundary pooling (Sec.~\ref{sec:refinement}). The boundary features are exploited together with the feature pyramid to output a fine-grained prediction $\{(\Delta\hat{\psi}_i, \Delta\hat{\xi}_i, \hat{y}_i^R)\}$ for both temporal regression and action classification. 

\subsection{Basic Prediction Module \label{sec:basic}}
We first build a basic anchor-free prediction module to get coarse temporal boundaries. For instance, for the feature of the $l$-th FPN level $f^l\in \mathbb{R}^{T_l\times C}$, we first project it to features $f^l_{loc}$ and $f^l_{cls}$ embedded in two latent space corresponding to localization and classification by two branches with two temporal convolutions respectively. 
Both of these two features $f^l_{loc}$ and $f^l_{cls}$ are processed with one layer of temporal convolution shared among all FPN layers to get coarse start and end boundary distances $(\hat{d}_i^s, \hat{d}_i^e)$ and class score $\hat{y}_i$ for each location $i$. Next we can get start and end time for $i$-th time step in $l$-th level as follow: 
\begin{equation}
    \begin{aligned}
    \hat{\psi}_i &=i*2^l - \hat{d}_i^s, \\
    \hat{\xi}_i &=i*2^l + \hat{d}_i^e.
\end{aligned}
\end{equation}
$T_l$ proposals are generated for $l$-th FPN layer in all. Such a simple framework can already temporally detect actions in a anchor-free manner, which enjoys several merits including no requirement for pre-defined anchors and less but more accurate predictions as discussed in Sec.~\ref{sec:1}. In the following sections, we focus on designing appropriate modules and training strategy for anchor-free TAL methods with better performance.

\subsection{Saliency-based Refinement Module \label{sec:refinement}}
As mentioned in Sec.~\ref{sec:1}, several existing works have shown the importance of boundary feature in TAL, especially for predicting the temporal distance. However, since different action instances could have various lengths, it is hard to attain these boundary information for all proposals via several simple temporal convolutions because of the limited receptive field. Therefore, we propose a saliency-based refinement module which is illustrated in Fig.~\ref{fig:refinement}(a), where we utilize the FPN features along with the coarse proposals to help our model gather boundary features to refine the predictions. For simplicity, we omit the subscript standing for FPN layers in the following detail. 
Take the localization feature $f_{loc}$ for example, first we project it into two latent spaces sensitive to start and end activities respectively via convolutional layers:
\begin{equation}
    \begin{aligned}
    f^s &= \sigma(\textrm{GN}(\text{Conv1}(f_{loc}))) \quad \in \mathbb{R}^{T_l\times C}, \\
    f^e &= \sigma(\textrm{GN}(\text{Conv2}(f_{loc})))\quad \in \mathbb{R}^{T_l\times C},
    \label{eq:start}
\end{aligned}
\end{equation}
where $\sigma$ and $\textrm{GN}$ denote ReLU and Group Normalization~\cite{wu2018group}. With the projection, the model can learn start and end sensitive signals separately, thus leaving less learning load to the FPN features for better training.

Then, given the coarse boundary results $\{(\hat{\psi}_k, \hat{\xi}_k)\}_{k=1}^{T_l}$ of the corresponding $l$-th FPN level, the $k$-th start and end regions $T_s^k$, $T_e^k$ are constructed as:
\begin{equation}
    \begin{aligned}
    T_s^k &= \left[\hat{\psi}_k-\frac{\hat{w}_k}{\delta_a},\hat{\psi}_k+\frac{\hat{w}_k}{\delta_b}\right], \\
    T_e^k &= \left[\hat{\xi}_k-\frac{\hat{w}_k}{\delta_b},\hat{\xi}_k+\frac{\hat{w}_k}{\delta_a}\right],
    \label{eq:region}
\end{aligned}
\end{equation}
where $\hat{w}_k=\hat{\psi}_k-\hat{\xi}_k$ means the length of proposal, $\delta_a, \delta_b$ are hyper-parameters controlling the proportion of selected regions outside and inside the proposal.

Next, an aggregation function $\mathcal{A}$ is applied to $f^s$ and $f^e$ in start and end regions respectively to collect the relevant boundary features.
Despite a lot of instantiations of $\mathcal{A}$ in former works, such as mean pooling~\cite{gao2017turn}, Gaussian weighted average~\cite{long2019gaussian} and directly gathering and concatenating~\cite{lin2020dbg}, these methods would introduce useless knowledge from frames not representing the action boundaries, thus blocking the model from precise predictions. Hence we propose a novel boundary pooling method to get the moment-level boundary feature $\hat{f}^s, \hat{f}^e\in \mathbb{R}^{T_l\times C}$ as shown in Fig.~\ref{fig:refinement}(b). The boundary pooling works as following:
\begin{equation}
    \begin{aligned}
    \hat{f}^s(k, i)&=\max_{j\in T_s^k}f^s(j, i)\quad  i=1,\cdots,C, \\
    \hat{f}^e(k, i)&=\max_{j\in T_e^k}f^e(j, i)\quad  i=1,\cdots,C.
\end{aligned}
\label{eq:pooling}
\end{equation}
Maximization is utilized aiming to select the largest activated cell, \textit{i.e.}, the most salient moment, for each channel along the temporal region. Note that as FPN goes deeper, the temporal dimension decreases to be too small for boundary pooling to find the appropriate boundary. Therefore, we add a shared frame-level feature $f_{frame}$ by applying upsample and several convolutions to the bottom FPN feature, from which start and end frame-level features $\tilde{f}^s, \tilde{f}^e$ are extracted for each proposal with the same projection and pooling procedure as in Eq.~\ref{eq:start} and Eq.~\ref{eq:pooling}.

After boundary pooling, the refined features are built by concatenating original features and all boundary features. A temporal convolution is applied to reduce channels:
\begin{equation}
    \hat{f}=\text{Conv}(f || \hat{f}^s || \hat{f}^e || \tilde{f}^s || \tilde{f}^e),
\end{equation}
where $\cdot||\cdot$ denotes channel-wise concatenation. These features are again employed as input of a simple network with temporal convolution to predict offsets of regression  $(\Delta\hat{\psi}, \Delta\hat{\xi})$, which can be added to the coarse predictions to get a fine-grained ones $(\tilde{\psi}, \tilde{\xi})$, and refined class score $\hat{y}^R$.

\subsection{Boundary Consistency Learning {\label{sec:BCL}}}
Although the boundary pooling can extract the most salient features, it cannot guarantee that the pooled features represent the true action boundary. Such a property is pivotal since if boundary pooling focuses on a background frame, the model cannot have enough useful information, and thus being misled to wrong results. To regularize our boundary pooling, we further propose Boundary Consistency Learning (BCL), which has two components: Activation Guided Learning and Boundary Contrastive Learning.

\noindent\textbf{Activation Guided Learning {\label{sec:AGL}}} Specifically, we re-scale the sensitive features $f^s$ and $f^e$  and take channel-wise mean:
\begin{equation}
    \tilde{g}^s=\mathrm{mean}(\tanh(f^s)), \tilde{g}^e=\mathrm{mean}(\tanh(f^e)).
\end{equation}
These two features can be seen as confidence revealing the probability of occurrence of start or end moment. We can obtain the ground truth $g^s,g^e \in \mathbb{R}^T$ by following definition:
\begin{equation}
    \begin{aligned}
g^s(i) &=\mathbb{I}(i\in \mathcal{B}(\psi_m) \text{ for } \forall m\in[1, M_X]), \\
g^e(i) &=\mathbb{I}(i\in \mathcal{B}(\xi_m) \text{ for } \forall m\in[1, M_X]),
\end{aligned}
\end{equation}
where $\mathcal{B}(\cdot)$ denotes the neighbor and $\mathbb{I}(\cdot)$ is the indicator function. After that we can calculate the Cross Entropy:
\begin{equation}
    \ell_{act}=\text{BCE}(g^s, \tilde{g}^s)+\text{BCE}(g^e, \tilde{g}^e),
    \label{eq:act}
\end{equation}
where BCE denotes the binary cross entropy. With $g^s$ and $g^e$ as guidance, we can constrain the feature to have high activation at the occurrence and closure of each action.

\noindent\textbf{Boundary Contrastive Learning} Consider a video $X$ containing an action instance $A$ and other background. If we split the action and fill in a random part of background, we will have two incomplete action fragments $A_1, A_2$ and background $Bg$ between them. Applying boundary pooling to these three regions leads to three pairs of features $(f^s_{A_1}, f^e_{A_1}),(f^s_{A_2}, f^e_{A_2}),(f^s_{Bg}, f^e_{Bg})$. In general, since $A_1$ and $A_2$ are continuous, the sensitive features $f^e_{A_1}$ and $f^s_{A_2}$ should be similar to each other and distant to $f^s_{Bg}$ and $f^e_{Bg}$ if we restrict boundary pooling to only make use of few frames inside the actions (\textit{i.e.}, a large $\delta_b$ in Eq.~\ref{eq:region}). However, this property could be broken when model is learnt to take high activations on background. In such situation, $f^e_{A_1}$ would be close to $f^s_{Bg}$ and $f^s_{A_2}$ would be close to $f^e_{Bg}$. Therefore, a good way to guarantee appropriate features is to apply the contrastive learning on these features, enlarging the distance between the sensitive features of action fragments and background. Formally, it can be realized by utilizing the following triplet objective function:
\begin{equation}
    \ell_{trip}=\max (\|f^e_{A_1}-f^s_{A_2}\|^2-\|f^e_{A_1}-f_{Bg}\|^2+1, 0),
    \label{eq:trip}
\end{equation}
where $f_{Bg}\in \left\{f_{Bg}^s, f_{Bg}^e\right\}$. In practice, we first count the minimal action length $w_{min}$ in one video. Next if the video has an action instance whose length is larger than $2\cdot w_{min}$ and a background clip with length $w_{min}$, we include this video into our contrastive learning pool and implement the above splitting procedure. In this way, both split action and background are long enough to be distinguished. In together, our Boundary Consistency Learning can be summarized into the following form:
\begin{equation}
    \ell_{con}=\ell_{act}+\ell_{trip}.
    \label{eq:con}
\end{equation}

\subsection{Training and Inference}
\noindent\textbf{Label Assignment}
For coarse prediction, we assign each location $i$ as a positive sample to ground truth $j$ when $\psi_j \leq i \leq \xi_j$ is satisfied. For refined prediction, we calculate the temporal IoU  (tIoU) between each coarse boundary prediction and the corresponding ground truth. A location $i$ is regarded as positive if its tIoU score is greater than 0.5. Denote $N_C, N_R$ as the number of positive samples for coarse and refined predictions individually.

\noindent\textbf{Training} Once having both the coarse and refined prediction of temporal boundary and class label, we can optimize the model with the following objective function:
\begin{equation}
    \mathcal{L}=\ell^C_{cls}+\lambda \ell^C_{loc}+\ell^R_{cls}+\lambda \ell^R_{loc}+\gamma \ell_{q},
    \label{eq:det}
\end{equation}
where $\lambda, \gamma$ are hyper-parameters, $\ell^C_{cls}, \ell^R_{cls}$ are softmax focal loss $\ell_{focal}$~\cite{lin2017focal} between both classification prediction $\left\{\hat{y}^C, \hat{y}^R\right\}$ and ground truth labels $y$:
\begin{equation}
    \ell_{cls}(\{\hat{y}_i\})=\frac{1}{N}\sum_{i}\ell_{focal}(\hat{y}_i, y_i),
\end{equation}
where $N\in \{N_C, N_R\}$. $\ell^C_{loc}$ is a tIoU loss between coarse boundaries $\hat{\phi}_i=(\hat{\psi}_i, \hat{\xi}_i)$ and the corresponding ground truth $\phi_i=(\psi_i, \xi_i)$. $\ell^R_{loc}$ is a L1 loss between the predicted offset $\hat{\Delta}_i=(\Delta\hat{\psi}_i, \Delta\hat{\xi}_i)$ and the corresponding offset label $\Delta_i=(\Delta\psi_i, \Delta\xi_i)$:
\begin{equation}
    \begin{aligned}
    \ell^C_{loc}(\{\hat{\phi}_i\}) &= \frac{1}{N_C}\sum_i\mathbb{I}(y_i \geq 1)(1 - \frac{|\hat{\phi}_i \cap \phi_i|}{|\hat{\phi}_i \cup \phi_i|}),\\
    \ell^R_{loc}(\{\hat{\Delta}_i\}) &= \frac{1}{N_R}\sum_i\mathbb{I}(y_i \geq 1)(|\hat{\Delta}_i - \Delta_i|).
\end{aligned}
\end{equation}
$\ell_{q}$ is a quality loss used to suppress the proposals with low quality. As a counterpart in the object detection, FCOS~\cite{tian2019fcos} proposes the centerness of each spatial location as the quality target. However, such definition of centerness for actions is vague since it is hard to decide the exact frame being a start or end signal of an action, and thus it is inappropriate to directly use centerness in TAL. To better predict quality of proposals, we utilize tIoU between boundary prediction $\tilde{\phi}$ and the location labels as the learning target of quality confidence $\eta$ generated from refined feature $\hat{f}$:  
\begin{equation}
    \ell_{q}(\{\eta_i\}) = \frac{1}{N_R}\sum_i\mathbb{I}(y_i \geq 1)\text{BCE}(\eta_i, \frac{|\tilde{\phi}_i \cap \phi_i|}{|\tilde{\phi}_i \cup \phi_i|}).
    \label{eq:quality}
\end{equation}
For each batch in training, we first optimize the model with $\mathcal{L}$. Then we seek the data available for BCL in Sec.~\ref{sec:BCL} and train the model with $\ell_{con}$ in Eq.~\ref{eq:con}.

\noindent\textbf{Inference}
For the $i$-th temporal location in $l$-th FPN layer, the final predictions are formalized through all outputs from our model, including coarse predictions $\hat{\psi}_{l,i}, \hat{\xi}_{l,i}, \hat{y}^C_{l,i}$ and refined ones $\Delta\hat{\psi}_{l,i}, \Delta\hat{\xi}_{l,i}, \hat{y}^R_{l,i}, \eta_{l,i}$, in the following form:
\begin{equation}
    \begin{aligned}
    \hat{w}_{l,i} &= \hat{\xi}_{l,i}-\hat{\psi}_{l,i}, \\
    \tilde{\psi}_{l,i} &= \hat{\psi}_{l,i}+\frac{1}{2}\hat{w}_{l,i}\Delta\hat{\psi}_{l,i}, \\
    \tilde{\xi}_{l,i} &= \hat{\xi}_{l,i}+\frac{1}{2}\hat{w}_{l,i}\Delta\hat{\xi}_{l,i}, \\
    \hat{y}_{l,i} &= \frac{1}{2}(\hat{y}^C_{l,i}+\hat{y}^R_{l,i})\eta_{l,i}.
\end{aligned}
\end{equation}
We then assemble all predictions and process them with Soft-NMS~\cite{bodla2017soft} to suppress redundant proposals.

\section{Experiments}
\begin{table*}[t]
 \centering
 \small
 {
  \begin{tabular}{ l|l|c|cccccc|cccc}
  \hline
 \multirow{2}{*}{Type} & \multirow{2}{*}{Model} & \multirow{2}{*}{Backbone} & \multicolumn{6}{c}{THUMOS14} & \multicolumn{4}{c}{ActivityNet1.3}\tabularnewline
 \cline{4-13}
& & & 0.3 & 0.4 & 0.5 & 0.6  & 0.7 & Avg. & 0.5 & 0.75 & 0.95 & Avg.\\

   \shline
    \multirow{6}{*}{Anchor-based} & SSAD~\cite{lin2017ssad} & TS & 43.0 & 35.0 & 24.6 & --- & --- & --- & --- & --- & --- & --- \tabularnewline \cline{4-13}
   & TURN~\cite{gao2017turn} & C3D & 44.1 & 34.9 & 25.6 & --- & --- & --- & --- & --- & --- & --- \tabularnewline \cline{4-13}
   & R-C3D~\cite{xu2017rc3d} & C3D & 44.8 & 35.6 & 28.9 & --- & --- & --- & 26.8 & --- & --- & --- \tabularnewline \cline{4-13}
   & CBR~\cite{gao2017cbr} & TS & 50.1 & 41.3 & 31.0 & 19.1 & 9.9 & 30.3 & --- & --- & --- & --- \tabularnewline \cline{4-13}
   & TAL~\cite{chao2018tal} & I3D & 53.2 & 48.5 & 42.8 & 33.8 & 20.8 & 39.8 & 38.2 & 18.3 & 1.3  & 20.2\tabularnewline \cline{4-13}
   & GTAN~\cite{long2019gaussian} & P3D & 57.8 & 47.2 & 38.8 & --- & --- & ---  & \textbf{52.6} & 34.1 & 8.9 & 34.3 \tabularnewline \cline{4-13}
    \hline
     \multirow{7}{*}{Actionness} & CDC~\cite{shou2017cdc} & --- & 40.1 & 29.4 & 23.3 & 13.1 & 7.9 & 22.8   & 45.3 & 26.0 & 0.2 & 23.8 \tabularnewline \cline{4-13}
        & SSN~\cite{zhao2017ssn} & TS & 51.0 & 41.0 & 29.8 & --- & --- & --- & 43.2 & 28.7 & 5.6 & 28.3 \tabularnewline \cline{4-13}
   & BSN~\cite{lin2018bsn} & TS & 53.5 & 45.0 & 36.9 & 28.4 & 20.0 & 36.8 & 46.5 & 30.0 & 8.0 & 30.0  \tabularnewline \cline{4-13}
   & BMN~\cite{lin2019bmn} & TS & 56.0 & 47.4 & 38.8 & 29.7 & 20.5 & 38.5 & 50.1 & 34.8 & 8.3 & 33.9 \tabularnewline \cline{4-13}
   & DBG~\cite{lin2020dbg} & TS & 57.8 & 49.4 & 42.8 & 33.8 & 21.7 & 41.1  & --- & --- & --- & ---\tabularnewline  \cline{4-13}
   & G-TAD~\cite{xu2020gtad} & TS & 54.5 & 47.6 & 40.2 & 30.8 & 23.4 & 39.3 & 50.4 & 34.6 & 9.0 & 34.1   \tabularnewline \cline{4-13}
   & BU-TAL~\cite{zhao2020bottom} & I3D & 53.9 & 50.7 & 45.4 & 38.0 & 28.5 & 43.3 & 43.5 & 33.9 & 9.2 & 30.1 \tabularnewline \cline{4-13}
   & BC-GNN~\cite{bai2020boundary} & TS & 57.1 & 49.1 & 40.4 & 31.2 & 23.1 & 40.2 & 50.6 & 34.8 & \textbf{9.4} & 34.3 \tabularnewline \cline{4-13}
    \hline
     \multirow{3}{*}{Other} & A2Net~\cite{yang2020a2net} & I3D & 58.6 & 54.1 & 45.5 & 32.5 & 17.2 & 41.6 & 43.6 & 28.7 & 3.7 & 27.8 \tabularnewline \cline{4-13}
        & G-TAD+PGCN~\cite{zeng2019pgcn} & I3D & 66.4 & 60.4 & 51.6 & 37.6 & 22.9 & 47.8 & --- & --- & --- & ---  \tabularnewline \cline{4-13}
    \hline
  Anchor-free  & Ours & I3D & \textbf{67.3} & \textbf{62.4} & \textbf{55.5} & \textbf{43.7} & \textbf{31.1} & \textbf{52.0} & 52.4 & \textbf{35.3} & 6.5 & \textbf{34.4} \tabularnewline
     \hline
  \end{tabular}
 }
 \vspace{0.05in}
\caption{\label{tab:result_thumos} Performance comparison with state-of-the-art methods on THUMOS14 and ActivityNet1.3, measured by \textit{m}AP at different IoU thresholds, and average \textit{m}AP in $[0.3:0.1:0.7]$ on THUMOS14 and $[0.5:0.05:0.95]$ on ActivityNet1.3.}

\vspace{-0.1in}
\end{table*}
\subsection{Datasets and Settings}
\noindent\textbf{Datasets} To validate the efficacy of our model, we conduct extensive experiments on commonly-used benchmark THUMOS14~\cite{jiang2014thumos} and ActivityNet1.3~\cite{caba2015activitynet}. \textit{THUMOS14} is composed of 200 validation videos and 212 testing videos from 20 categories labeled for temporal localization. \textit{ActivityNet1.3} has 19,994 videos with 200 action classes. We follow the former setting~\cite{lin2018bsn} to split this dataset into training, validation and testing subset by 2:1:1.

\noindent\textbf{Implementation Details}
On THUMOS14, we sample both RGB and optical flow frames at 10 frames per second (fps) and split video into clips. The length of each clip $T$ is set as 256 frames. Adjacent clips have a temporal overlap of $m$ frames and $m$ is set to 30 in training and 128 in testing. On ActivityNet1.3, we sample frames using different fps and ensure the number of video frames is 768 for each video. Thus, each video has only one clip with 768 frames. On both datasets, the frame spatial size is set to $96 \times 96$. Random crop, horizontal flipping are used as data augmentation in training. To extract features of clips, we finetune a I3D~\cite{carreira2017quo} model pre-trained on Kinetics.

Our model is trained for 16 epoches using Adam~\cite{kingma2014adam} with learning rate of $10^{-5}$, weight decay of $10^{-3}$. Batch size is set to 1. We set $\delta_a$ to 4 and $\delta_b$ to 100 for $\ell_{con}$ and $\delta_b$ to 10 for other loss terms. The weight of loss $\lambda$ is set to 10 on THUMOS14 and 1 on ActivityNet1.3 and $\gamma$ is set to 1 empirically. In the testing phase, the results of RGB and optical flow frames are averaged to obtain final locations and class scores. The tIoU threshold in Soft-NMS is set to 0.5 for THUMOS14 and 0.85 for ActivityNet1.3. 

\noindent\textbf{Metrics} We report mean Average Precision (\textit{m}AP) in all experiments. The thresholds are $[0.3:0.1:0.7]$ for THUMOS14 and $[0.5:0.05:0.95]$ for ActivityNet1.3.

\subsection{Main Results}
We compare our model with state-of-the-art methods in Tab.~\ref{tab:result_thumos} and report the backbone used by each method, including TS~\cite{simonyan2014two}, C3D~\cite{tran2015learning}, P3D~\cite{qiu2017learning} and I3D~\cite{carreira2017quo}. On THUMOS14, our AFSD outperforms the strongest competitor A2Net and G-TAD on all thresholds by large margin, especially 7.7\% on \textit{m}AP@0.6. The remarkable improvement comes along with high efficiency, thus making our model more practical for real TAL scenarios. Note that while A2Net also has an anchor-free module, the performance of their merged model is far worse than ours, not to mention the sole anchor-free branch, which proves the superiority of our architecture for anchor-free methods. 

On ActivityNet1.3, the results are still comparable. Our model receives the best \textit{m}AP@0.75 and average \textit{m}AP compared to the strongest competitor GTAN. It is noteworthy that while all of the actionness-guided methods have less average \textit{m}AP than ours, most of them enjoy a higher \textit{m}AP@0.95. One possible reason is that they can enumerate all potential proposals, thus the ground truth proposals are already contained in the alternative prediction set. With such an enumeration strategy the actionness-based methods would be better when dealing with harder datasets like ActivityNet1.3. Compared with actionness-guided methods, our model is more efficacious in the sense of producing less proposals and attaining better overall performance when considering multiple thresholds.  Additionally, compared with THUMOS14, ActivityNet is less well annotated as explained by official report \cite{alwassel2018diagnosing} showing ‘it is hard to agree about the temporal boundaries’, which partially attributes to the slight improvement.


\subsection{Ablation Study}
To further verify the efficacy of our contributions, we conduct several ablation studies on THUMOS14 for the RGB model, including each part of our model and the choice of hyper-parameters.

\begin{table*}[t]
\centering
\begin{minipage}{0.3\textwidth} 
\centering 
{
  \setlength{\tabcolsep}{0.6mm}{
  \begin{tabular}{l|cccc}
  \hline
 Model&  0.5 & 0.6  & 0.7 & Avg.\\
   \shline
    baseline & 43.1 & 31.0 & 19.0 & 40.4 \tabularnewline
    +centerness & 43.3 & 31.6 & 17.7 & 40.2 \tabularnewline 
    +quality & \textbf{44.0} & \textbf{32.0} & \textbf{19.8} & \textbf{41.4} \tabularnewline 
     \hline
  \end{tabular}}
}
\end{minipage}
\quad
\begin{minipage}{0.3\textwidth} 
\centering 
{
  \setlength{\tabcolsep}{0.6mm}{
  \begin{tabular}{l|cccc}
  \hline
 Model&  0.5 & 0.6  & 0.7 & Avg.\\

   \shline
     0-1 clip & 45.3 & 34.6 & 22.4 & 42.6 \tabularnewline 
     0-1 norm & 45.4 & 34.9 & 21.6 & 42.9 \tabularnewline 
     \textit{tanh} & \textbf{45.9} & \textbf{35.0} & \textbf{23.4} & \textbf{43.5}\tabularnewline 
     \hline
  \end{tabular}}}
\end{minipage}
\quad
\begin{minipage}{0.3\textwidth} 
\centering 
{
  \setlength{\tabcolsep}{1.3mm}{
  \begin{tabular}{l|cccc}
  \hline
 Model& 0.5 & 0.6  & 0.7 & Avg.\\

   \shline
    $\delta_a = \delta_b$ & 45.0 & 33.4 & 21.2 & 42.3\tabularnewline 
    $\delta_a > \delta_b$ & 45.3 & 34.7 & 21.8 & 42.6\tabularnewline 
    $\delta_a < \delta_b$ & \textbf{45.9} & \textbf{35.0} & \textbf{23.4} & \textbf{43.5} \tabularnewline 
     \hline
  \end{tabular}}}
\end{minipage}
\vspace{0.05in}
\\
\quad
\begin{minipage}{0.3\textwidth} \raggedright
(a) \textbf{Training strategy}: we compare our quality confidence with centerness loss proposed in FCOS.
\end{minipage}
\quad
\begin{minipage}{0.3\textwidth} \raggedright
(b) \textbf{Feature normalization}: we compare different choice of feature normalization in constraint.
\end{minipage}
\quad
\begin{minipage}{0.3\textwidth} \raggedright
(c) \textbf{Boundary feature extraction}: we compare different choice of boundary regions.
\end{minipage}
\vspace{0.15in}
\\
\begin{minipage}{0.3\textwidth} 
\centering 
{
  \setlength{\tabcolsep}{1.35mm}{
  \begin{tabular}{l|cccc}
  \hline
 Model&  0.5 & 0.6  & 0.7 & Avg.\\
   \shline
    naive & 44.9 & 32.5 & 19.9 & 41.6\tabularnewline
    self & 44.6 & 33.9 & 22.3 & 42.4 \tabularnewline 
    frame & 42.9 & 31.9 & 20.2 & 40.5 \tabularnewline 
    all & \textbf{45.9} & \textbf{35.0} & \textbf{23.4} & \textbf{43.5} \tabularnewline 
     \hline
  \end{tabular}}
}
\end{minipage}
\quad
\begin{minipage}{0.3\textwidth} 
\centering 
{
  \setlength{\tabcolsep}{1.1mm}{
  \begin{tabular}{l|cccc}
  \hline
 Model&  0.5 & 0.6  & 0.7 & Avg.\\

   \shline
    mean & 45.1 & 34.6 & 22.1 & 42.7 \tabularnewline 
    conv & 44.8 & 34.4 & 22.3 & 42.6 \tabularnewline
    stack & 44.9 & 33.6 & 22.3 & 42.9 \tabularnewline
    max & \textbf{45.9} & \textbf{35.0} & \textbf{23.4} & \textbf{43.5} \tabularnewline 
     \hline
  \end{tabular}}}
\end{minipage}
\quad
\begin{minipage}{0.3\textwidth} 
\centering 
{
  \setlength{\tabcolsep}{0.6mm}{
  \begin{tabular}{l|cccc}
  \hline
 Model&  0.5 & 0.6  & 0.7 & Avg.\\

   \shline
    w/o BCL & 44.3 & 34.1 & 21.2 & 42.0\tabularnewline 
    $\ell_{act}$ & 45.6 & 34.4 & 22.3 & 42.7 \tabularnewline
    $\ell_{trip}$ & 45.5 & 34.8 & 22.4 & 42.7 \tabularnewline
    $\ell_{act}+\ell_{trip}$ & \textbf{45.9} & \textbf{35.0} & \textbf{23.4} & \textbf{43.5} \tabularnewline 
     \hline
  \end{tabular}}}
\end{minipage}
\vspace{0.05in}
\\
\quad
\begin{minipage}{0.3\textwidth} \raggedright
(d) \textbf{Refinement}: we compare different sources of information for refinement.
\end{minipage}
\quad
\begin{minipage}{0.3\textwidth} \raggedright
(e) \textbf{Instantiations}: we compare different forms of boundary feature extraction.
\end{minipage}
\quad
\begin{minipage}{0.3\textwidth} \raggedright
(f) \textbf{Consistency learning}: we compare models varied with Boundary Consistency Learning.
\end{minipage}
\vspace{0.15in}
\\

\caption{\label{tab:ablation} Ablation studies of RGB model on THUMOS14, measured by \textit{m}AP at 0.5, 0.6 and 0.7, and average \textit{m}AP in $[0.3:0.1:0.7]$.}
\vspace{-0.2in}
\end{table*}

\begin{table}[t]
 \centering
 {
  \begin{tabular}{ l|c|c}
  \hline
 Model & GPU &FPS\tabularnewline
   \shline
   S-CNN~\cite{shou2016scnn} & --- & 60\tabularnewline \cline{3-3}
   DAP~\cite{escorcia2016daps} & --- & 134\tabularnewline \cline{3-3}
   CDC~\cite{shou2017cdc} & TITAN Xm & 500\tabularnewline \cline{3-3}
   SS-TAD~\cite{buch2019end} & TITAN Xm & 701\tabularnewline \cline{3-3}
   R-C3D~\cite{xu2017rc3d} & TITAN Xm & 569\tabularnewline \cline{3-3}
   R-C3D~\cite{xu2017rc3d} & TITAN Xp & 1030\tabularnewline \cline{3-3}
   Ours & 1080 Ti & \textbf{3259}\tabularnewline \cline{3-3}
   Ours & V100 & \textbf{4057}\tabularnewline 
     \hline
  \end{tabular}
 }
 \vspace{0.05in}
\caption{\label{tab:time} Comparison of inference speed between our method and other methods on THUMOS14.}

\vspace{-0.2in}
\end{table}

\noindent\textbf{Effectiveness of Quality Confidence} We first compare the basic prediction module introduced in Sec.~\ref{sec:basic} trained and inferred with and without the proposed quality confidence in Tab.~\ref{tab:ablation}(a). Besides, we add an extra model using the centerness proposed in FCOS. The results show that centerness leads to 1.3\% drop on \textit{m}AP@0.7, doing no help for training TAL model. In contrast, model with quality loss $\ell_q$ (Eq.~\ref{eq:quality}) can have 1.0\% average \textit{m}AP improvement, which supports our claim that the definition of centerness is somehow inappropriate in TAL, thus cannot be directly applied. Compared with that, our quality loss $\ell_q$ is a more suitable objective function for suppressing low-quality action proposals.

\noindent\textbf{Choice of Signal Normalization} In Action Guided Learning introduced in Sec.~\ref{sec:AGL}, we adopt a \textit{tanh} function to normalize the feature vector. We compare this one with another two instantiations. One is a hard clipping between $\left[0,1\right]$. The other is to use simple standardization in the following form, which is denoted as \textbf{0-1 norm} in Tab.~\ref{tab:ablation}(b):
\begin{equation}
    f(i)=\frac{f(i)-\min_{i=1}^T f(i)}{\max_{i=1}^Tf(i)-\min_{i=1}^T f(i)}.
\end{equation}
The results show that using \textit{tanh} can have 0.9\% and 0.6\% average \textit{m}AP advantage against two alternatives.

\noindent\textbf{Choice of Boundary Region} In Tab.~\ref{tab:ablation}(c) we assess choices of the boundary region used for pooling boundary features, which is composed of (1) a symmetric area with the coarse boundary as center point, denoted as $\delta_a=\delta_b$. (2, 3) two asymmetric regions either focusing on background or action. Through the results we can safely say that it is better to keep a larger proportion of region inside the coarse action than the background. The possible reason is that our naive predictor can already produce a relatively accurate prediction. As shown in Tab.~\ref{tab:ablation}(a), the performance of baseline can beat most of the competitors in Tab.~\ref{tab:result_thumos} even if the competitors fuse RGB and optical flow models for final prediction and our baseline only utilizes RGB frames as input.

\noindent\textbf{Effectiveness of Boundary Refinement} In Tab.~\ref{tab:ablation}(d) we compare four forms of boundary refinement, including: (1) \textbf{Naive}: Only several temporal convolutions are directly applied to $f_{loc}, f_{cls}$ to get another prediction. In this way, the information disposed for refinement is the stacked neighbor introduced by the convolutions. (2) \textbf{Self}: The saliency-based refinement module is utilized without $f_{frame}$, and thus the consistency learning only calculates losses corresponding to the FPN feature. (3) \textbf{Frame-level}: only $f_{frame}$ is adopted in temporal refinement and FPN features are abandoned. (4) \textbf{All}: All usable features are included, which is our final model. The results demonstrate that: (1) Boundary information is more valuable for refinement than that from neighborhood of each temporal location. (2) Frame-level feature can only be taken as a complement of FPN features. Using only frame-level feature would result in 1.1\% average \textit{m}AP drop against the naive refinement model. 


\noindent\textbf{Instantiation of Boundary Pooling} We compare our proposed boundary pooling with the following three instantiations: (1) \textbf{Mean}: The max operation is replaced with mean. (2) \textbf{Conv}: We sample three temporal positions for each region, and a 1-layer temporal convolution is applied to aggregate them. (3) \textbf{Stack}: Similar to (2), while instead of using convolution, we directly concatenate these three features into one boundary feature. Note that all models are trained with our boundary consistency learning. The results are presented in Tab.~\ref{tab:ablation}(e). Among all instantiations, pooling with max receives the best performance, showing 0.8\%, 0.9\% and 0.6\% advantage of average \textit{m}AP against mean, conv and stack respectively. It is noteworthy that the improvement of our model over others is larger than 1.0\% on \textit{m}AP@0.7, which indicates that moment-level boundary feature helps the model generate more accurate proposals.

\noindent\textbf{Effectiveness of Boundary Consistency Learning} We verify the proposed BCL by comparing our full model with that trained only with $\mathcal{L}$ (Eq.~\ref{eq:det}) and without $\ell_{con}$ (Eq.~\ref{eq:con}). Results in Tab.~\ref{tab:ablation}(f) suggest that when trained without any consistency guarantee, the model cannot learn good representations for boundary pooling. Therefore lack of useful information in refinement leads to worse performance. Moreover, each loss term brings 0.7\% improvement on average \textit{m}AP and by assembling $\ell_{act}$ (Eq.~\ref{eq:act}) and $\ell_{trip}$ (Eq.~\ref{eq:trip}) together, our model finally achieves the best performance. To further validate the efficacy of BCL, we obtain the learned frame-level feature $f_{frame}$ of an action instance and its neighboring background, each half part of which could represent the start and end signals in hypothesis. 

\noindent\textbf{Comparison on Inference Time}
As discussed, the proposed AFSD is highly efficient. To verify this claim we report the inference speed on THUMOS14 in terms if fps among different models in Tab.~\ref{tab:time}. Thus results show that our model is much more faster than the existing methods. The main reasons is that after features are extracted we can employ both a lighter predictor and a lighter refinement module composed of 1D convolution due the the help of anchor-free mechanism. Moreover,  the number of proposals produced by our model is less than that of these methods, which also helps speed our model.

\section{Conclusion}
In this paper we explore the possibility of a novel form of temporal action localization model --- anchor-free method. We discuss the merits over anchor-based methods, and actionness-guided methods and design a dedicated anchor-free model. Our model includes an end-to-end trainable basic predictor and a temporal refinement module. For the refinement module, we analyze the drawbacks of existing means to extract boundary features, and propose a novel boundary pooling together with a Boundary Consistency Learning strategy. We achieve remarkable results on THUMOS14 and comparable ones on ActivityNet1.3. The results indicate the strength of anchor-free model as a promising choice for solving temporal action localization.

{\small
\bibliographystyle{ieee_fullname}
\bibliography{egbib}
}

\end{document}